\RequirePackage{fix-cm}
\documentclass[twocolumn]{svjour3}          
\smartqed  
\usepackage{graphicx}
\usepackage[utf8]{inputenc}
\usepackage{amsmath}
\usepackage{amssymb}
\usepackage{booktabs}
\usepackage{multirow}
\usepackage{subcaption}
\usepackage{algorithm}
\usepackage{algpseudocode}
\usepackage{microtype}
\usepackage{inconsolata}
\usepackage{listings}
\usepackage{hyperref}

\makeatletter
\def\makeheadbox{}
\makeatother

\begin{document}

\title{A Hybrid Kinematic GCN and Elastic Transformer for Efficient Indian Sign Language Recognition}
\subtitle{}

\author{Pushp Garg \and Himangshu Sarma \and Mrinmoy Ghorai}

\institute{Human Computer Interaction Lab, IIIT SriCity, India \\
           \email{\{pushp.g23, himangshu.sarma\}@iiits.in, mgre04@gmail.com}
}

\date{ } 

\maketitle

\begin{abstract}
Sign Language Recognition (SLR) presents complex challenges due to variable signing speeds and spatial intricacies. Specifically, Indian Sign Language (ISL) requires precise modeling of its inherently bimanual nature and is bottlenecked by the scarcity of large-vocabulary benchmarks and the inability of standard models to handle irregular signing velocities. To address this, a structurally-aware hybrid framework for efficient, gloss ISL recognition is proposed. To minimize computational redundancy, an entropy-driven SVD sampler that dynamically extracts informative keyframes is used. Because this subsampling creates irregular physical time gaps, the proposed architecture employs a Continuous Dual Positional Encoding paired with a distance-aware Elastic Transformer to preserve natural signing rhythm. Spatially, a Kinematically-Conditioned Adaptive GCN explicitly models bimanual geometry using unified bone-joint features. To validate our framework, a 120-class, 9,453-video native ISL dataset is created. Augmented by Physical Node Dropout and ArcFace optimization to mitigate subject-specific memorization, our framework achieves a competitive 89.8\% accuracy, establishing a strong baseline for complex sign language modeling.

\end{abstract}

\section{Introduction}
\label{sec:intro}

Disabling hearing loss affects over 430 million people globally, a figure projected to reach 700 million by 2050 \cite{number_of_deaf}. For millions of individuals, sign language serves as the primary mode of communication \cite{sl}. Despite the existence of over 300 sign languages used by 70 million people \cite{300_sl_1,300_sl_2}, Indian Sign Language (ISL) remains severely under-resourced \cite{isl_resources_1}. ISL is a highly dynamic visual language featuring rapid micro-motions, prolonged hold-phases, and complex bimanual coordination \cite{isl_is_bimanual}, necessitating architectures with precise spatial and adaptive temporal modeling capabilities.

Standard spatial-temporal models, including Graph Convolutional Networks (GCNs) and Video Transformers, face critical bottlenecks in this domain \cite{arib2025signformer}. Uniform frame sampling wastes computation on static hold-phases while risking the loss of rapid kinematic transitions \cite{deng2025vtan}. Conversely, dynamically dropping redundant frames creates irregular physical time gaps that disrupt standard Transformers' sequential rhythm assumptions \cite{chen2023contiformer}.

To address these limitations, an efficient, structurally-aware framework for large-vocabulary ISL recognition is proposed. The proposed model is a hybrid architecture: a Kinematically-Conditioned Adaptive Part-GCN handles spatial skeleton structure via graph convolution, while a distance-aware Elastic Transformer performs temporal reasoning over the resulting pose tokens. This cross-architectural design bridges the complementary strengths of graph-based spatial modeling and attention-based temporal reasoning for irregular sequences. To minimize computational redundancy, our framework utilizes an entropy-driven dimensionality reduction pipeline based on Singular Value Decomposition (SVD) \cite{stewart1993early} and Kullback-Leibler (KL) divergence \cite{kullback1951information} to extract only the most kinematically innovative keyframes. To process these irregularly sampled frames without losing natural signing rhythm, a Continuous Dual Positional Encoding with a distance-aware Elastic Transformer is paired. Spatially, the Part-GCN processes a unified 8-channel spatio-kinematic embedding to accurately model bimanual geometry.

Finally, to mitigate subject-memorization and distinguish visually similar classes, Physical Node Dropout alongside an Additive Angular Margin Loss (ArcFace) \cite{arcface} is employed. To robustly validate our architecture, a novel 9,453-video, 120-class native ISL benchmark featuring diverse abstract terminology is created. Unlike existing ISL corpora that suffer from severe class sparsity and force models into few-shot paradigms, our dataset provides substantial depth (averaging $\sim$78 videos per gloss) with signer-disjoint train and test partitions. This density provides the necessary foundation to effectively learn identity-invariant kinematics.

In summary, our primary contributions are:
\begin{itemize}
    \item A new 120-class, 9,453-video native ISL dataset. On this benchmark, our model achieves an 89.8\% Test Accuracy and 90\% Macro F1-Score, demonstrating strong generalization with a minimal ($\sim$3.5\%) generalization gap.
    \item An SVD-driven cumulative subsampling module paired with Continuous Dual Positional Encodings, enabling the Elastic Transformer to process irregular time steps while preserving physical rhythm.
    \item Our hybrid framework, which efficiently bridges spatial graph convolutions and temporal transformers to model isolated sign language kinematics.
\end{itemize}

\section{Related Work}
\label{sec:related_work}
\paragraph{Broader Sign Language Processing}
The research community has actively pursued tools and techniques for processing sign languages, establishing an intersection between the computer vision \cite{li2020word} and natural language processing \cite{yin2021including} domains. Consequently, a wide array of sign language processing (SLP) tasks have been formalized. These include sign language detection \cite{moryossef2020real}, identification \cite{monteiro2016detecting}, segmentation \cite{bull2020automatic}, generation \cite{xiao2020skeleton,saunders2020progressive}, and continuous translation \cite{camgoz2018neural,muller2022findings,yin2020attention,muller2023considerations,jiang2023machine}. While self-supervised pretraining \cite{selvaraj2022openhands} and phonological modeling \cite{tavella2022wlasl} have recently gained traction, the severe unavailability of densely annotated resources remains the primary challenge for robust, data-driven SLP approaches.

\paragraph{ISL Datasets and Recognition Constraints}
Data-driven Sign Language Recognition (SLR) is strictly constrained by the availability of large-scale benchmarks. Early Indian Sign Language (ISL) research predominantly focused on static alphabet recognition using 2D object detectors \cite{swamy2025indian} or limited-vocabulary gestures \cite{mistree2020towards,odia,kirandevraj2025continuous}. For dynamic temporal glosses, large-vocabulary corpora (\textbf{Table \ref{tab:isl_datasets}}) like CISLR, introduced by \cite{isl_resources_2}, and INCLUDE \cite{include} provided foundational baselines. However, their severe sparsity (e.g., CISLR averaging 1.5 videos per word) forces models into complex few-shot or prototypical learning paradigms to compensate for data scarcity \cite{aziz2025attention}. While recent large-scale continuous benchmarks address sentence-level translation \cite{isl_resources_1,joshi2023isltranslate}, a gap remains for deep, isolated gloss evaluation where high sample density per class allows models to learn identity-invariant kinematics. Our proposed 120-class dataset directly addresses this by providing substantial depth (averaging $\sim$78 videos per class), delivering the data density necessary to learn identity-invariant kinematics without few-shot workarounds.

\paragraph{Architectural Modeling for Sign Language}
Standard spatial temporal architectures, including 3D CNNs and standard Video Transformers, struggle to capture the complex bimanual nature and irregular rhythmic velocities of ISL, resulting in sub-optimal recognition \cite{isl_resources_1,joshi2023isltranslate}. Furthermore, standard temporal graph networks implicitly assume uniform frame sampling \cite{zhou2022tgl}. This assumption is inherently flawed for sign language: it wastes computation on static hold-phases while risking the loss of rapid micro-motions. As demonstrated in our ablation study (Table~\ref{tab:ablation_study}), replacing entropy-driven SVD sampling with uniform temporal sampling causes a 2.3\% accuracy drop, directly confirming that content-aware frame selection is critical. The proposed architecture directly addresses these established bottlenecks by replacing naive spatial processing with a Kinematically-Conditioned Adaptive GCN and bypassing the uniform-sampling assumption via a distance-aware Elastic Temporal Transformer.

\paragraph{Irregular Temporal Modeling}
A parallel line of work in sequence modeling addresses the challenge of non-uniform temporal spacing. Continuous-time Transformers such as ContiFormer \cite{chen2023contiformer} extend standard attention to handle continuous-time inputs by incorporating time-delta information into the attention kernel, while Neural ODE-based approaches model latent dynamics between irregular observations. Our Continuous Dual Positional Encoding and distance-aware attention bias are motivated by this paradigm but are specifically tailored for sign language: rather than modeling continuous-time dynamics, we encode both the discrete logical ordering of entropy-selected keyframes and their absolute physical timestamps, allowing the Elastic Transformer to reason about irregular physical gaps without sacrificing sequential coherence.

\begin{table}[htbp]
\centering
\caption{Comparison of existing Indian Sign Language (ISL) datasets based on vocabulary size, total videos, and average videos per word.}
\label{tab:isl_datasets}
\resizebox{\columnwidth}{!}{%
\begin{tabular}{p{3cm}ccc}
\toprule
\textbf{Datasets} & \textbf{Words} & \textbf{Videos} & \textbf{Avg. Samples} \\
\midrule
\cite{nandy2010recognition} & 22 & 600 & 27.3 \\
\cite{kishore2012video} & 80 & 800 & 10 \\
INCLUDE \cite{include} & 263 & 4287 & 16.3 \\
CISLR \cite{isl_resources_2} & 4765 & 7050 & 1.5 \\
ISL-CSLTR \cite{isl_csltr} & 186 & 700 & 3.8 \\
\midrule
Our & 120 & 9453 & 78.8\\
\bottomrule
\end{tabular}
}
\end{table}

\begin{figure*}
    \begin{center}
        \includegraphics[width=0.9\textwidth]{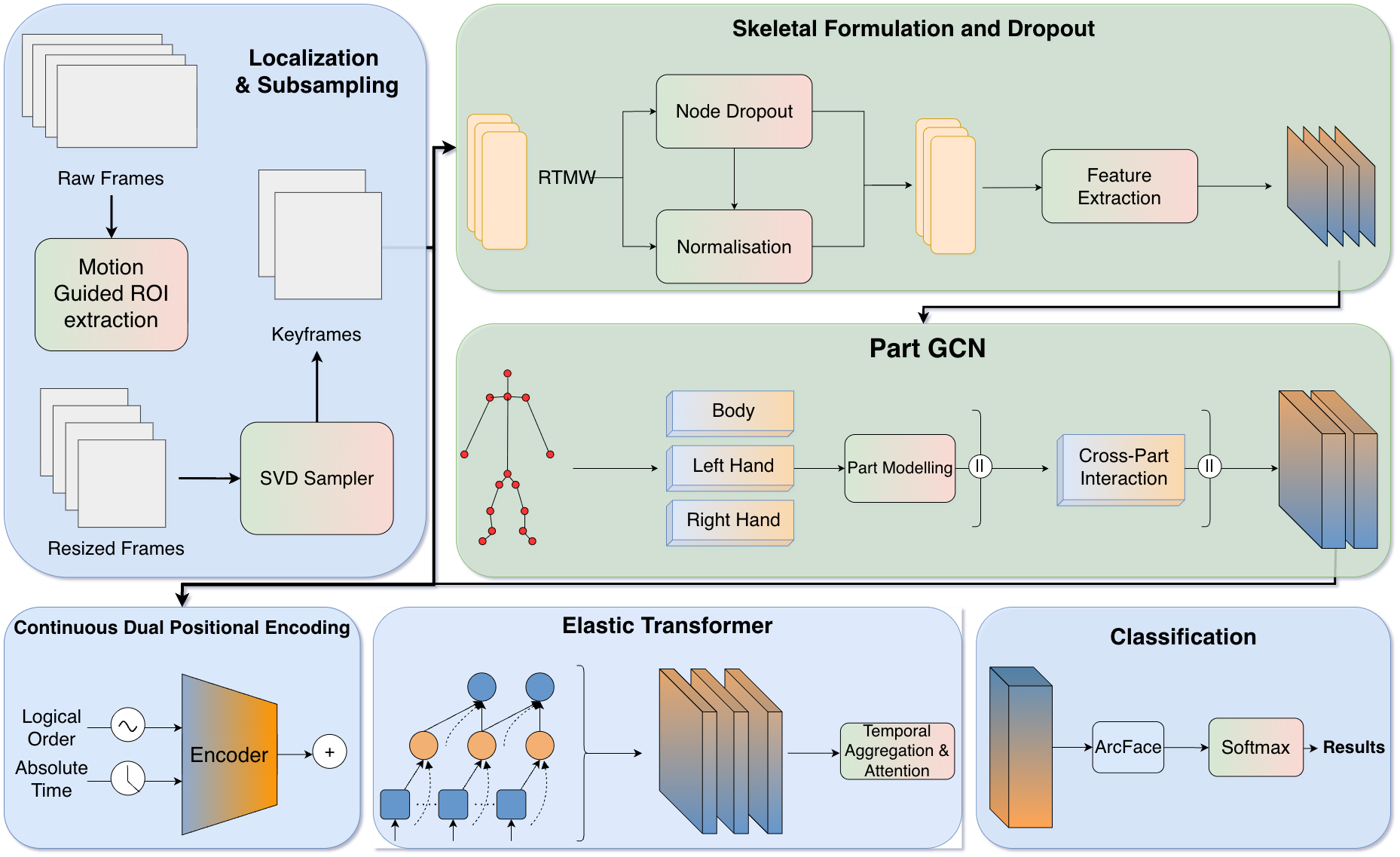}
    \end{center}
    \caption{Overall architecture of the proposed hybrid framework. The architecture is a hybrid: a Kinematically-Conditioned Part-GCN handles spatial skeleton structure, while the distance-aware Elastic Transformer performs temporal reasoning over the resulting pose tokens.}
    \label{fig:architecture}
\end{figure*}

\section{Method}
\label{sec:methods}
The proposed hybrid framework (Figure \ref{fig:architecture}) transforms raw video into structurally-aware skeletal tokens through dynamic localization, entropy-driven subsampling, and spatio-temporal modeling.

\subsection{Localization and Subsampling}
To minimize background interference, a Motion-Guided ROI extraction module computes a global motion heatmap. Given a video $V = \{F_1, \dots, F_n\}$, temporal innovation between Gaussian-smoothed frames is $\Delta_t(x, y) = |(F_t \ast G) - (F_{t-1} \ast G)|$. Motion energy is accumulated into a global map $A$ by thresholding pixel-wise differences ($\tau_L$):
\begin{equation}
    A(x, y) = \sum_{t=2}^{n} \mathbb{I}(\Delta_t(x, y) > \tau_L)
\end{equation}
Otsu's thresholding \cite{otsu} generates a binary mask $S$, followed by morphological closing $S_{closed} = (S \oplus K) \ominus K$, where $\oplus$ and $\ominus$ denote dilation and erosion by structuring element $K$, respectively. This operation fills interior holes and smooths contour boundaries to extract a clean, localized square bounding box.

To preserve signing dynamics while pruning redundant hold-phases, a Cumulative SVD-Innovation Sampler extracts grayscale proxies $P_t$, performs SVD ($P_t=U \Sigma V^T$), and normalizes the singular value spectrum:
\begin{equation}
    s_t(i) = \frac{\sigma_i}{\sum_{j} \sigma_j + \epsilon}
\end{equation}
A sequence-specific threshold $\Theta$ is defined using Kullback–Leibler (KL) Divergence modified by a sparsity factor $\alpha$:
\begin{equation}
\begin{split}
    \Theta &= \mu(D_{KL}(s_t \parallel s_{t-1})) \\
    &\quad + \alpha \cdot \sigma(D_{KL}(s_t \parallel s_{t-1}))
\end{split}
\end{equation}
Crucially, a frame $t$ is selected as a keyframe only if its innovation relative to the last selected keyframe $k$ exceeds $\Theta$:
\begin{equation}
    D_{KL}(s_t \parallel s_k) > \Theta
\end{equation}
This enforces a minimum stride ($t - k > 1$) and acts as a lightweight gatekeeper, drastically reducing the number of frames passed to the heavy pose extractor.

\subsection{Skeletal Formulation and Dropout}
Selected keyframes are processed via RTMPose-Wholebody \cite{rtmpose} to extract a 65-node topology ($V=65$), yielding coordinates $C \in \mathbb{R}^{T \times V \times d}$. To mitigate subject-memorization, Physical Node Dropout randomly occludes an entire subgraph (Left Hand, Right Hand, or Body) during training with probability $p$:
\begin{equation}
    \hat{C}_{t, i} = C_{t, i} \odot M_i \quad ; \quad M_i = 0 \ \forall i \in \mathcal{V}_{subgraph}
\end{equation}
Surviving coordinates are translation-normalized via the nose centroid. An 8-channel spatio-kinematic tensor $X_{in} \in \mathbb{R}^{T \times V \times 8}$ is then formulated from five biomechanically-motivated feature groups: spatial coordinates $(X,Y)$ provide absolute joint position; velocity $v_t = (\hat{C}_t - \hat{C}_{t-1}) / \max(\Delta t, \epsilon)$ encodes dynamic motion; spatial bone vectors $\vec{b}_i = \hat{C}_i - \hat{C}_{anchor}$ capture limb orientation; joint angles $\cos(\theta_{i, j}) = (\vec{b}_i \cdot \vec{b}_j) / (\|\vec{b}_i\| \|\vec{b}_j\| + \epsilon)$ represent relative joint configuration invariant to subject scale; and pose confidence scores down-weight unreliable landmarks during propagation.

\subsection{Kinematically-Conditioned Adaptive Part-GCN}
To ground the network in human biomechanics, spatial processing is initialized from a binary physical adjacency matrix prior $A_{prior} \in \{0,1\}^{V \times V}$, where $A_{prior,ij} = 1$ if joints $i$ and $j$ share a direct anatomical connection and 0 otherwise. The learnable adjacency matrix $A_{learn}$ is initialized as $A_{prior} + \mathcal{N}(0, 0.01)$. The graph convolution operation employs a row-wise Softmax to dynamically normalize edges and allow asymmetric directed information flow:
\begin{equation}
    Z = \text{Softmax}(A_{learn}, \text{dim}=-1) (X_{in} W) + b
\end{equation}

Instead of uniform processing, the architecture partitions the 65-node spatio-kinematic tensor into three independent subgraphs Body (23 nodes), Left Hand (21 nodes), and Right Hand (21 nodes) each processed by a dedicated Part-GCN block to extract robust, localized kinematic representations.

To capture the bimanual coordination crucial to sign language, the network computes Euclidean distances between physical anchors (e.g., wrist-to-wrist, index-to-index, hands-to-face). These distances are embedded via a Multi-Layer Perceptron (MLP) to form a cross-part interaction token. Finally, localized outputs from the three Part-GCNs are concatenated with this interaction token, constructing a holistic, structurally-aware pose token.

\subsection{Continuous Dual PE and Elastic Transformer}
To capture localized micro-motions prior to self-attention, the sequentially embedded tokens $X_{pe} \in \mathbb{R}^{B \times D_{pose} \times T}$ pass through a 1D Temporal Convolutional Network (TCN) within a residual block:
\begin{equation}
    X_{local} = \mathcal{F}_{TCN}(X_{pe}) + X_{pe}
\end{equation}

Because entropy-driven subsampling creates irregular time gaps, standard discrete positional encodings fail. Input sequences are therefore mapped to a Continuous Dual Positional Encoding. A learnable embedding matrix $E_{rel} \in \mathbb{R}^{T_{max} \times (D/2)}$ captures the discrete, logical sequence order of the surviving keyframes, while absolute physical timestamps $t_{abs} \in [0,1]$ drive sinusoidal encodings \cite{vaswani2017attention}:
\begin{align}
 PE_{abs}(t_{abs}, 2i) &= \sin\left(\frac{t_{abs}}{10000^{2i / (D/2)}}\right) \\
 PE_{abs}(t_{abs}, 2i+1) &= \cos\left(\frac{t_{abs}}{10000^{2i / (D/2)}}\right)
\end{align}

Global reasoning is performed by Elastic Transformer Blocks. A Distance-Aware Attention Bias dynamically modulates the attention matrix based on the physical temporal gap $\delta_{ij} = |t_{abs, i} - t_{abs, j}|$ between tokens $i$ and $j$, using a multi-layer perceptron:
\begin{equation}
\begin{split}
    \text{Attention}&(Q, K, V) = \\
    &\text{Softmax}\left(\frac{QK^T}{\sqrt{d_k}} + \text{MLP}(\delta_{ij}) + \mathcal{M}\right)V
\end{split}
\end{equation}
A Masked Temporal Aggregator condenses the variable-length sequence into a video-level vector $v$ by learning scalar attention weights $\alpha_t$ over all $T$ time steps:
\begin{align}
    \alpha_t &= \text{Softmax}_t\!\left(\mathbf{w}^T \tanh(W_a X_{local, t} + b_a)\right) \\
    v &= \sum_{t=1}^{T} \alpha_t X_{local, t}
\end{align}
where $\text{Softmax}_t$ denotes normalization over the time dimension, and $X_{local,t}$ is the post-TCN, post-Transformer token at step $t$.

\subsection{Angular Margin Optimization}
\label{arcface}
Standard linear classification is replaced with an Additive Angular Margin Loss (ArcFace) \cite{arcface}. ArcFace enforces a fixed angular margin $m$ between classes on a hypersphere, producing more compact intra-class and more discriminative inter-class feature distributions than standard cross-entropy a property particularly beneficial for sign language, where many glosses share overlapping sub-movements. Given L2-normalized representations $v$ and classifier weights $W_j$:
\begin{equation}
\label{eq:arcface}
    \mathcal{L}_{ArcFace} = -\log \frac{e^{s \cdot \cos(\theta_{y_i} + m)}}{e^{s \cdot \cos(\theta_{y_i} + m)} + \sum_{j \neq y_i} e^{s \cdot \cos \theta_j}}
\end{equation}
To equitably propagate gradients across the dataset's inherent class imbalance, this objective is weighted by the inverse class frequency $w_{y_i} \propto 1/\text{freq}(y_i)$. The final training loss is therefore:
\begin{equation}
    \mathcal{L}_{Total} = -\sum_{i=1}^{N} w_{y_i} \log \left( p_{y_i}^{(\text{ArcFace})} \right)
\end{equation}
where $p_{y_i}^{(\text{ArcFace})}$ is the scaled softmax probability derived from Eq.~\eqref{eq:arcface}.

\section{Experiments}
\label{sec:Experiments}
This section comprehensively evaluates the efficacy and efficiency of the proposed framework. A 120-class native ISL benchmark is introduced. Subsequently, the architectural contributions are evaluated through rigorous comparisons against established baselines and public datasets, concluding with detailed ablation studies and computational profiling.

\subsection{Dataset Description}
\label{sec:dataset}
The scarcity of large-vocabulary skeletal benchmarks severely bottlenecks dynamic ISL recognition. To provide a robust testbed, a native 120-class ISL repository comprising 9,453 videos (averaging 42.2 frames per video) is collected. The vocabulary spans 5 semantic categories (\textbf{Table \ref{tab:dataset_distribution}}). Class frequencies range from 33 to 97 videos per gloss; this imbalance necessitates the dynamic class-weighted loss formulation from Section \ref{arcface}.

\begin{table}[ht]
\centering
\caption{Dataset distribution across semantic categories including word counts, weight ranges, and signer averages.}
\label{tab:dataset_distribution}
\begin{tabular}{lccc}
\toprule
\textbf{Category} & \textbf{Words} & \textbf{Class Weight} & \textbf{Avg. samples} \\
\midrule
Medical & 27 & [0.64, 1.26] & 61.30 \\
Geographical & 23 & [0.90, 2.66] & 86.90 \\
Technical & 18 & [0.90, 2.66] & 87.11 \\
Action Verbs & 20 & [0.86, 2.52] & 86.25 \\
Daily Life & 32 & [0.85, 2.47] & 78.30 \\
\midrule
\textbf{Full Dataset} & \textbf{120} & \textbf{[0.81, 2.41]} & \textbf{78.80} \\
\bottomrule
\end{tabular}
\end{table}

\paragraph{Acquisition and Partitioning:}
Data was recorded using a Kinect V2 at 15 FPS. To maximize real-world webcam generalizability, the RGB stream is deliberately utilized, explicitly ignoring depth and infrared modalities. The dataset is partitioned via a class-stratified split into a training repository (7,527 samples) and a held-out test repository (1,926 samples) used for all final evaluations. To support reproducibility and future research, the dataset and model code will be made publicly available upon acceptance of this work.

\begin{figure*}[t]
    \begin{center}
        \includegraphics[width=0.9\textwidth]{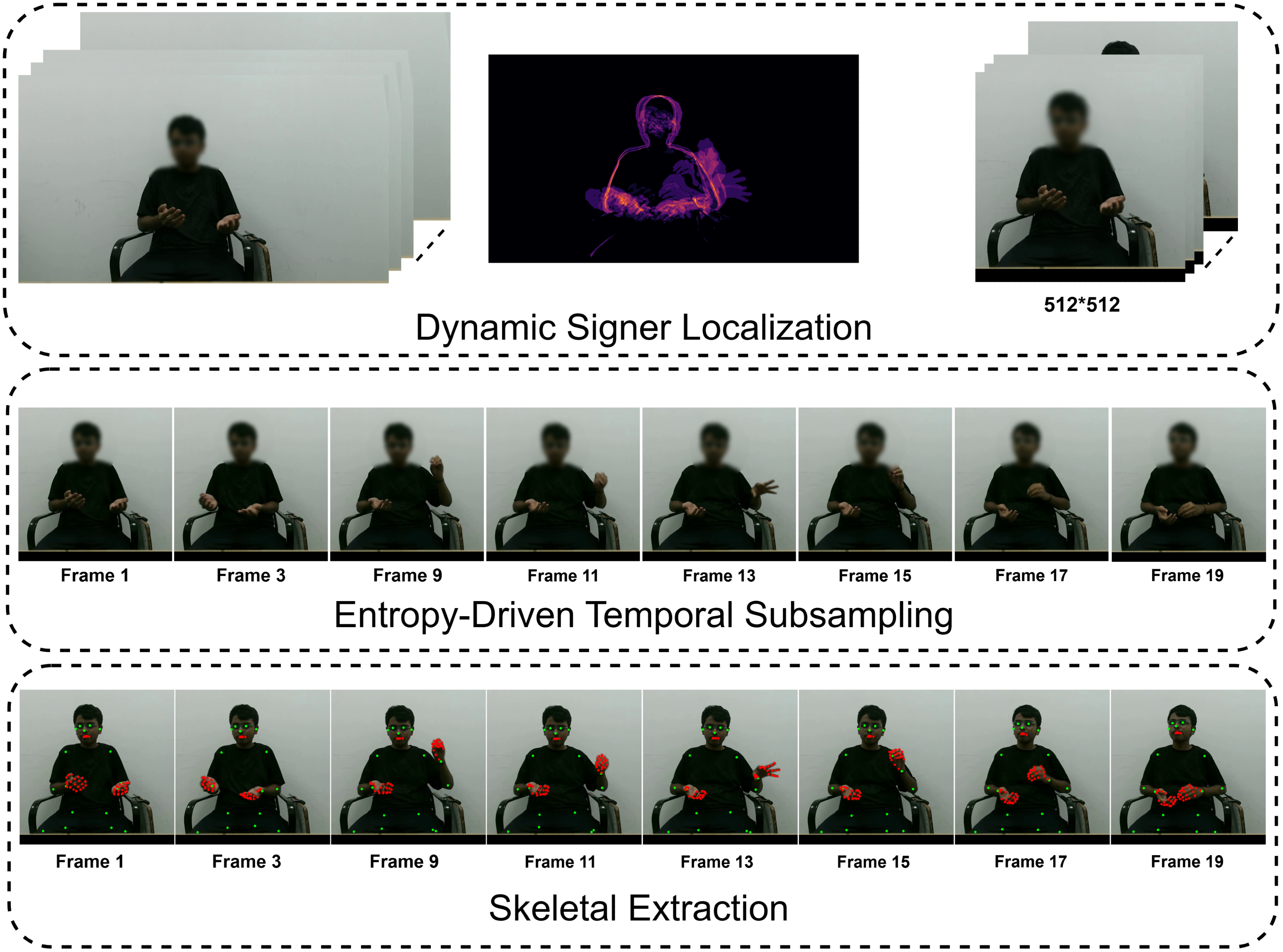}
    \end{center}
    \caption{The dynamic preprocessing pipeline of our framework. Raw sequences undergo Motion-Guided Region of Interest (ROI) extraction, followed by entropy-driven SVD temporal subsampling, and finally RTMW skeletal extraction to form the unified input representation.}
    \label{fig:preprocessing}
\end{figure*}

\paragraph{Signer Distribution and Identity Protocol:}
Data was collected over several years through multiple acquisition drives at academic institutions and community schools, yielding hundreds of distinct volunteer signers. Crucially, each volunteer performed any given gloss at most once. Consequently, for every gloss class, the set of signers present in the training partition is strictly disjoint from the set present in the test partition guaranteeing a zero-shot identity scenario for each word. While the total count of unique signers across all classes was not formally tracked due to the distributed, multi-year nature of collection, the scale and diversity of acquisition sites make cross-partition identity overlap across classes negligibly unlikely. This protocol forces the architecture to learn invariant kinematic semantics rather than memorizing subject-specific anatomy.

\subsection{Data Preprocessing}
\label{sec:preprocessing}

To maximize computational efficiency, the raw dataset is dynamically processed (\textbf{as illustrated in Figure \ref{fig:preprocessing}}) before entering the main architecture.

\paragraph{Visual Localization:} Raw videos are first localized via Motion-Guided ROI extraction using a temporal innovation threshold $\tau_L$ and padding factor $\lambda$. The active signing region is resized to a $512 \times 512$ grayscale proxy to mitigate background and lighting variances.

\paragraph{Dynamic SVD Subsampling:} To eliminate temporal redundancy and variable signing velocities, the SVD-Innovation Sampler extracts high-entropy keyframes based on a sparsity multiplier $\alpha$ and sequence constraint $T_{max}$. This pipeline successfully compresses the raw input from an average of 42.2 frames down to a sparse 13.2 keyframes per video a 68.6\% reduction in temporal dimensionality that significantly minimizes computational overhead without sacrificing critical gestural semantics.

\paragraph{Input Feature Composition:} The extracted keyframes undergo skeletal pruning via RTMW to isolate a 65-node topology tailored for ISL: 23 upper-body/facial nodes ($\mathcal{V}_{body}$) and 21 nodes per hand ($\mathcal{V}_{lh}, \mathcal{V}_{rh}$). The coordinates are normalized relative to the nose centroid and shoulder width. Finally, they are expanded into the unified 8-channel spatio-kinematic representation, forming the final input tensor $X_{in} \in \mathbb{R}^{T \times 65 \times 8}$.

\subsection{Training Details and Reproducibility}
\label{sec:training}

Experiments utilized an Intel i9-13900K CPU and a 24GB RTX 4090 GPU. While temporal subsampling used grayscale proxies, RTMPose skeletal extraction utilized full-resolution RGB for maximum landmark precision. The network trained for 100 epochs using AdamW (with high weight decay to regularize the transformer) and a ReduceLROnPlateau scheduler. Strict deterministic behavior ($seed=42$) across splits and augmentations guaranteed exact reproducibility.

From the 7,527 training samples, an internal class-stratified 80/20 split yields approximately 6,022 samples for gradient updates and 1,505 samples for an internal validation subset, used exclusively for checkpoint selection via minimum validation loss monitoring. All reported performance metrics are derived strictly from the separate, held-out test repository of 1,926 samples, which is never used during training or model selection. All exhaustive configuration parameters including spatial dimensions, ArcFace margins, and learning rates are consolidated in \textbf{Appendix \ref{sec:hyperparameter}}.

\paragraph{Quantitative Performance}
Our framework is evaluated on our native 120-class dataset, deriving all metrics strictly from the held-out test partition using the checkpoint with the minimum validation loss. As detailed in \textbf{Table \ref{tab:category_analysis}}, the model achieves an 89.8\% Test Accuracy and a 90.06\% Macro F1-Score. This tight alignment indicates that our dynamic class-weighting and ArcFace angular margin successfully mitigate inherent dataset imbalance, ensuring equitable feature representation across both frequent and rare classes.

\begin{table}[ht]
\centering
\caption{Category-level performance breakdown of the proposed model.}
\label{tab:category_analysis}
\resizebox{\columnwidth}{!}{%
\begin{tabular}{lcccc}
\toprule
\textbf{Category} & \textbf{Train Acc} & \textbf{Train F1} & \textbf{Test Acc} & \textbf{Test F1} \\
\midrule
Medical & 99.2\% & 99.2\% & 98.2\% & 98.2\% \\
Geographical & 97.0\% & 96.8\% & 91.0\% & 90.1\% \\
Technical & 98.0\% & 98.0\% & 90.7\% & 90.5\% \\
Action Verbs & 97.5\% & 97.5\% & 88.3\% & 88.3\% \\
Daily Life & 96.2\% & 96.3\% & 89.1\% & 89.2\% \\
\midrule
\textbf{Full} & \textbf{93.3\%} & \textbf{93.5\%} & \textbf{89.8\%} & \textbf{90.0\%} \\
\bottomrule
\end{tabular}
}
\end{table}

\paragraph{Training Dynamics and Generalization}
Over a 100-epoch schedule, the network demonstrated stable convergence. Optimal generalized weights were recovered at Epoch 80, achieving a minimal test loss of 2.088 alongside a Train Accuracy of 93.84\% and Train F1-Score of 94.14\%. The narrow $\sim$3.5\% generalization gap validates our structural regularization strategy. Consistent with our ablation results (Table~\ref{tab:ablation_study}), this resistance to overfitting is primarily attributable to Physical Node Dropout ($p=0.20$) and the Entropy-Driven Subsampling, which together force the model to learn invariant kinematic semantics rather than redundant, easily overfit spatial trajectories.

\begin{table*}[t!]
\centering
\caption{Performance of the proposed framework on established public sign language benchmarks. Training accuracies are reported for our model only to demonstrate the absence of underfitting; they were not available in the original baseline publications (denoted --).}
\label{tab:public_benchmarks}
\begin{tabular}{p{7cm}lcc}
\toprule
\textbf{Model} & \textbf{Dataset} & \textbf{Train Acc} & \textbf{Test Acc} \\
\midrule
MobileNetV2 BiLSTM \cite{include} & INCLUDE50 & -- & 94.5\% \\
DCDW-LSTM \cite{dey2023gesture} & INCLUDE50 & -- & 91.3\% \\
BERT Transformer \cite{prathap2023islr} & INCLUDE50 & -- & 89.5\% \\
C3D-BiLSTM MHAttention \cite{dey2024recognition} & INCLUDE50 & -- & 92.46\% \\
Meta-Learning + LSTM RNN \cite{katti2023character} & INCLUDE50 & -- & 84.52\% \\
MediaPipe CNN BiLSTM \cite{ghanimi2024open} & INCLUDE50 & -- & 97.44\% \\
Mediapipe Attention-based BiLSTM \cite{sujatha2024improved} & INCLUDE50 & -- & 88.63\% \\
MediaPipe LSTM \cite{khartheesvar2024automatic} & INCLUDE50 & -- & 94.8\% \\
EfficientNet MultiFeature Attention GRU \cite{yenisari2025deep} & INCLUDE50 & -- & 98.41\% \\
MobileNetV2 BiLSTM \cite{include} & INCLUDE & -- & 85.6\% \\
Transfer Learning Deep Learning \cite{mahyoub2023sign} & INCLUDE & -- & 96.0\% \\
InceptionV3-GRU \cite{marais2022investigating} & LSA64(Leave Two Subject Out) & -- & 74.22\% \\
Resnet-18 with RReLU \cite{akdaug2024isolated} & LSA64(Leave Two Subject Out) & -- & 94.68\% \\
BoW+SubCls \cite{ronchetti2016sign} & LSA64(Leave One Subject Out) & -- & 91.7\% \\
Cumulative SD-VLAD with SVM \cite{rodriguez2018towards} & LSA64(Leave One Subject Out) & -- & 85\% \\
Transformer \cite{alyami2024isolated} & LSA64(Leave One Subject Out) & -- & 91.09\% \\
LSTM \cite{konstantinidis2018sign} & LSA64(Random Split) & -- & 98.09\% \\
VGG-16, LSTM \cite{konstantinidis2018deep} & LSA64(Random Split) & -- & 99.84\% \\
MEMP network \cite{zhang2019dynamic} & LSA64(Random Split) & -- & 99.063\% \\
CNN, kernel-based extreme learning machine \cite{imran2020deep} & LSA64(Random Split) & -- & 97.81\% \\
\midrule
Proposed Framework (Ours) & INCLUDE50 & 99.3\% & 97.1\% \\
Proposed Framework (Ours) & INCLUDE & 98.2\% & 95.8\% \\
Proposed Framework (Ours) & LSA64(Leave Two Subject Out) & 98.5\% & 94.4\% \\
Proposed Framework (Ours) & LSA64(Stratified Sampling) & 98.6\% & 99.1\% \\
\bottomrule
\end{tabular}
\end{table*}

\subsection{Evaluation on Public Benchmarks}
\label{sec:public_eval}

To validate cross-domain and signer-independent generalization, our framework is evaluated on the Argentinian Sign Language \cite{lsa64} LSA64, INCLUDE50, and INCLUDE ISL benchmarks. For LSA64, a strict zero-identity-leakage protocol (Subjects 1--8 for training, 9--10 for testing) is applied.

Our SVD sampler demonstrated robust efficiency across framerates: on LSA64 (60 FPS), it condensed 127.8-frame sequences to 38.8 keyframes (69.6\% reduction); on INCLUDE configurations (25 FPS), 64.25-frame sequences were compressed to 26.7 keyframes (58.4\% reduction). This yields substantial computational savings during pose extraction while preserving critical kinematics. As detailed in \textbf{Table \ref{tab:public_benchmarks}}, our approach achieves competitive predictive performance on both datasets, proving our architecture successfully transfers generalized representations to unseen signers and diverse recording environments.

\subsection{Comparison with Baseline Architectures}
\label{sec:baseline_comparison}

To isolate our architectural contributions, standard skeleton baselines (ST-GCN, CTR-GCN) are evaluated on our 120-class benchmark using the exact same SVD-extracted spatio-kinematic features.

As detailed in \textbf{Table \ref{tab:baseline_comparison}}, our model significantly outperforms both baselines. The sharp performance drop in the standard networks (e.g., CTR-GCN dropping to 64.8\% Test Acc) highlights a fundamental architectural limitation: standard temporal graph networks implicitly assume uniform time steps and fail to interpret irregular temporal gaps. By replacing standard convolutions with a distance-aware Elastic Transformer and continuous positional encodings, the proposed architecture natively preserves physical signing rhythm, maximizing the utility of subsampled data.

\begin{table}[ht]
\centering
\caption{Comparison of our proposed model against established skeleton baselines on the 120-class ISL dataset. All models utilize the proposed SVD preprocessing pipeline.}
\label{tab:baseline_comparison}
\resizebox{\columnwidth}{!}{%
\begin{tabular}{lcc}
\toprule
\textbf{Architecture} & \textbf{Train Acc} & \textbf{Test Acc}\\
\midrule
CTR-GCN \cite{ctrgcn} & 69.9\% & 64.8\%\\
ST-GCN \cite{stgcn} & 85.1\% & 83.0\%\\
\midrule
\textbf{Proposed (Ours)} & \textbf{93.3\%}& \textbf{89.8\%}\\
\bottomrule
\end{tabular}
}
\end{table}

\begin{table*}[htbp]
\centering
\caption{Performance comparison of alternative vision architectures. These models were evaluated exclusively on the Geographical word category subset due to their extreme computational overhead.}
\label{tab:geo_comparison}
\begin{tabular}{llcc}
\toprule
\textbf{Approach} & \textbf{Pre-trained Prior} & \textbf{Train Acc (\%)} & \textbf{Test Acc (\%)} \\
\midrule
SlowFast\_R50 & Kinetics-400 & 63.40 & 30.62 \\
MViTv2 \cite{mvitv2} & Kinetics-400 & 33.84 & 33.63 \\
VideoMAE & Kinetics-400 & 99.00 & 42.60 \\
\midrule
VideoMAE \cite{videomae} & SSv2 & 99.00 & 74.62 \\
VideoMAE (Staged Fine-Tuning) & SSv2 & 100.00 & 81.20 \\
VideoMAE (LoRA Only) & SSv2 & 80.00 & 60.00 \\
VideoMAE (LoRA + Unfreeze) & SSv2 & 99.00 & 76.60 \\
V-JEPA-2 \cite{vjepa2} & SSv2 & 97.46 & 88.32 \\
\bottomrule
\end{tabular}
\end{table*}

\begin{table*}[t]
\centering
\caption{Ablation study of the proposed framework on the 120-class ISL dataset.}
\label{tab:ablation_study}
\begin{tabular}{lcccc}
\toprule
\textbf{Modification} & \textbf{Train Acc} & \textbf{Train F1} & \textbf{Test Acc} & \textbf{Test F1} \\
\midrule
\textbf{Proposed Full Framework (Ours)} & \textbf{93.3\%} & \textbf{93.5\%} & \textbf{89.8\%} & \textbf{90.0\%} \\
\midrule
\multicolumn{5}{@{}l}{\textit{Loss Functions}} \\
\quad w/o Weighted Loss & 93.1\% & 93.3\% & 89.4\% & 89.5\% \\
\quad w/o ArcFace & 94.2\% & 94.4\% & 88.0\% & 88.2\% \\
\quad w/o ArcFace + w/o Weighted Loss & 93.9\% & 94.2\% & 87.4\% & 87.8\% \\
\midrule
\multicolumn{5}{@{}l}{\textit{Architecture Modules}} \\
\quad w/o Absolute Positional Encoding (PE) & 93.9\% & 94.0\% & 88.5\% & 88.7\% \\
\quad w/o Physical Node Dropout & 96.3\% & 96.5\% & 88.1\% & 88.3\% \\
\quad w/o Temporal Convolution (TCN) & 91.2\% & 91.5\% & 88.0\% & 88.1\% \\
\midrule
\multicolumn{5}{@{}l}{\textit{Data \& Preprocessing}} \\
\quad w/o SVD Sampler (All Frames) & 96.0\% & 96.2\% & 91.0\% & 91.0\% \\
\quad Uniform Temporal Sampling & 91.5\% & 91.7\% & 87.5\% & 87.7\% \\
\quad w/o Kinematic Features (X,Y only) & 82.3\% & 83.0\% & 83.2\% & 83.0\% \\
\quad w/o PE + w/o Kinematic Features & 81.3\% & 82.0\% & 80.3\% & 80.4\% \\
\bottomrule
\end{tabular}
\end{table*}

Furthermore, \textbf{Table \ref{tab:geo_comparison}} presents a targeted comparison against highly parameterized vision architectures (VideoMAE, MViTv2). Even against these resource-intensive models, our lightweight framework demonstrates highly competitive performance.

\paragraph{Ablation Study}
To quantify the impact of our architectural contributions, an exhaustive ablation study (\textbf{Table \ref{tab:ablation_study}}) is conducted. Removing the Continuous Positional Encoding (PE) and Kinematic features causes a severe performance collapse (dropping to 80.3\% Test Acc), proving that standard sequence modeling fails on irregular temporal gaps without explicit spatial-temporal priors. The ablation also directly confirms the benefit of entropy-driven selection over uniform sampling (87.5\% vs. 89.8\%), validating that content-aware keyframe extraction is essential. Furthermore, bypassing the SVD Sampler to process ``All Frames'' yields only a marginal 1.2\% accuracy gain, validating that our dimensionality reduction pipeline successfully discards redundancy while fully preserving critical gestural semantics.

\subsection{Computational Efficiency and Hardware Deployment}
As demonstrated in our ablation study (Table \ref{tab:ablation_study}), processing the full raw sequence (``All Frames'') yields a marginal accuracy improvement (91.0\% vs 89.8\%). However, this requires processing an average of 42.2 frames per video. By utilizing the proposed entropy-driven SVD sampler, our framework reduces the temporal dimensionality by 68.6\% (averaging 13.2 keyframes), trading a 1.2\% accuracy drop for substantial computational savings.

Profiling the full proposed architecture (inclusive of the TCN and Transformer blocks) with a maximum sequence length of $T=25$ and $D_{pose}=512$ yields the highly efficient footprint detailed in \textbf{Table \ref{tab:efficiency}}.

\begin{table}[ht]
\centering
\caption{Computational footprint of the proposed architecture.}
\label{tab:efficiency}
\begin{tabular}{@{}ll@{}}
\toprule
\textbf{Metric} & \textbf{Measurement} \\
\midrule
Trainable Parameters & 6.13 Million \\
Theoretical Compute & 252.57 MFLOPs \\
Model Weights (float32) & 23.39 MB \\
Peak VRAM ($B=1$) & 34.91 MB \\
Inference Latency & 2.87 $\pm$ 0.19 ms \\
Throughput (RTX 4090) & 348.7 samples/sec \\
\bottomrule
\end{tabular}
\end{table}

This optimized profile is a direct consequence of the SVD sampler: because the self-attention mechanism's complexity scales quadratically with sequence length ($O(T^2)$), reducing the sequence to a maximum bound of $T=25$ (averaging 13.2) collapses the attention matrix footprint, allowing the 6.1M parameters to operate with high efficiency.

\section{Conclusion}
\label{sec:conclusion}
This work presented an efficient hybrid framework for large-vocabulary Indian Sign Language recognition. By unifying spatial graph convolutions with elastic temporal attention, the architecture effectively isolates invariant kinematic structures from raw video sequences. A cornerstone of this success is the dimensionality reduction pipeline; the entropy-driven temporal subsampling dynamically filters out redundant frames, significantly reducing computational overhead while preserving crucial gestural semantics. Combined with ArcFace optimization to enforce robust decision boundaries, the model achieved 89.8\% Test Accuracy and a 90.06\% Macro F1-score on a novel 120-class ISL benchmark, demonstrating its capability in handling complex, abstract gloss terminology.

\section*{Limitations}
\label{sec:limitations}
While our approach establishes a strong computational baseline, its reliance on pose extractors means severe occlusions or extreme motion blur can corrupt initial skeletal tokens and propagate errors through the network. Additionally, the primary dataset's acquisition at 15 FPS may restrict the recognition of rapid finger-spelling, despite the temporal convolution network's ability to capture local micro-motions. Furthermore, the current structural embedding focuses strictly on hand and body posture, lacking explicit tracking of non-manual markers such as fine-grained facial expressions or mouthing that are crucial to natural sign language linguistics. Future work will explore lightweight multimodal fusion to integrate these non-manual markers and expand the dimensionality reduction pipeline to handle dynamic frame-rate inputs. Finally, we plan to scale the dataset by collecting more isolated and continuous data, facilitating the evaluation of the framework on larger corpora and extending the architecture to sentence-level continuous translation tasks.

\bibliographystyle{spmpsci}      
\bibliography{reference}         

\appendix
\section{Hyperparameter Details}
\label{sec:hyperparameter}

\begin{table}[ht] 
    \centering
    \caption{\label{tab:all_hypers} Comprehensive summary of model implementation configuration and hyperparameters.}
    \begin{tabular}{@{}lc@{}} 
        \toprule
        \textbf{Parameter} & \textbf{Value} \\
        \midrule
        \multicolumn{2}{@{}l}{\textit{Hardware}} \\
        \quad GPU Model & RTX 4090 (24GB) \\
        \quad CPU Model & i9-13900K \\
        \midrule
        \multicolumn{2}{@{}l}{\textit{Preprocessing \& Split}} \\
        \quad Motion Threshold ($\tau_L$) & 25 \\
        \quad ROI Padding Factor ($\lambda$) & 1.2 \\
        \quad SVD Sparsity Factor ($\alpha$) & 0.1 \\
        \quad Max Sequence Length ($T_{max}$) & 25 \\
        \quad Train / Val Split Ratio & 80\% / 20\% \\
        \quad Random Seed & 42 \\
        \midrule
        \multicolumn{2}{@{}l}{\textit{Skeletal Augmentation}} \\
        \quad Unified Input Channels & 8 \\
        \quad Node Dropout Probability ($p$) & 0.20 \\
        \midrule
        \multicolumn{2}{@{}l}{\textit{Model Architecture}} \\
        \quad Spatial Subgraphs & Body, LH, RH \\
        \quad Transformer Layers ($N$) & 2 \\
        \quad Attention Heads ($H$) & 8 \\
        \quad TCN Kernel Size ($K$) & 3 \\
        \quad Pose Embedding Dim. ($D_{pose}$) & 512 \\
        \midrule
        \multicolumn{2}{@{}l}{\textit{ArcFace Optimization}} \\
        \quad Angular Margin ($m$) & 0.30 \\
        \quad Hypersphere Radius ($s$) & 30.0 \\
        \midrule
        \multicolumn{2}{@{}l}{\textit{Training (AdamW)}} \\
        \quad Base Learning Rate ($LR$) & $5 \times 10^{-4}$ \\
        \quad Batch Size & 32 \\
        \quad LR Scheduler Patience & 5 \\
        \quad Weight Decay & 0.5 \\
        \quad Total Epochs & 100 \\
        \bottomrule
    \end{tabular}
\end{table}

\end{document}